\documentclass[conference]{IEEEtran}
\IEEEoverridecommandlockouts
\usepackage{cite}
\usepackage{amsmath,amssymb,amsfonts}
\usepackage{algorithmic}
\usepackage{graphicx}
\usepackage{textcomp}
\usepackage{xcolor}
\usepackage{multirow}
\usepackage{booktabs}   
\usepackage{multirow}   
\usepackage{array}      
\usepackage{cleveref}

\renewcommand{\footnoterule}{%
  \kern -3pt
  \hrule width 0.2\textwidth height 0.5pt
  \kern 2pt
} 

\def\BibTeX{{\rm B\kern-.05em{\sc i\kern-.025em b}\kern-.08em
    T\kern-.1667em\lower.7ex\hbox{E}\kern-.125emX}}
\begin{document}

\title{A Deep Single Image Rectification Approach for Pan-Tilt-Zoom Cameras}

\author{
	Teng Xiao\textsuperscript{1,2},  Qi Hu\textsuperscript{1}, Qingsong Yan\textsuperscript{3}, Wei Liu\textsuperscript{*1,2}, Zhiwei Ye\textsuperscript{*1,2}, and Fei Deng\textsuperscript{3,4} \\
	\textsuperscript{1}School of Computer Science, Hubei University of Technology, Wuhan, China\\
    \textsuperscript{2}Hubei Key Laboratory of Green Intelligent Computing Power Network ,Wuhan, China\\
    \textsuperscript{3}School of Surveying and Mapping, Wuhan University, Wuhan, China\\
    \textsuperscript{4}Wuhan Tianjihang Information Technology Co., Ltd., Wuhan, China
}
\vspace{-5mm}

\maketitle
\renewcommand{\thefootnote}{}
\footnotetext{* Corresponding author. This work was funded by the National Natural Science Foundation of China (42301491, 62376089)}

\maketitle

\begin{abstract}
\label{sec:abstract}
Pan-Tilt-Zoom (PTZ) cameras with wide-angle lenses are widely used in surveillance but often require image rectification due to their inherent nonlinear distortions. 
Current deep learning approaches typically struggle to maintain fine-grained geometric details, resulting in inaccurate rectification. This paper presents a Forward Distortion and Backward Warping Network (FDBW-Net), a novel framework for wide-angle image rectification. It begins by using a forward distortion model to synthesize barrel-distorted images, reducing pixel redundancy and preventing blur. The network employs a pyramid context encoder with attention mechanisms to generate backward warping flows containing geometric details. Then, a multi-scale decoder is used to restore distorted features and output rectified images. FDBW-Net's performance is validated on diverse datasets: public benchmarks, AirSim-rendered PTZ camera imagery, and real-scene PTZ camera datasets. It demonstrates that FDBW-Net achieves SOTA performance in distortion rectification, boosting the adaptability of PTZ cameras for practical visual applications.
\end{abstract}


\section{Introduction}
\label{sec:intro}   
Pan-Tilt-Zoom (PTZ) cameras equipped with wide-angle lenses are widely used in surveillance due to their flexible remote control, which allows for effective monitoring of large areas. However, the inherent nonlinear optical distortion of wide-angle lenses presents significant challenges for image-based visual computing such as object localization\cite{jinlong2024wisecam} and scene understanding\cite{peng2023openscene}, making image rectification essential prior to application.

Traditional image rectification approaches have relied on multi-view geometry theories, estimating internal camera parameters from a set of multi-view images \cite{hartley2003multiple,xiao2024global}. 
However, these methods typically require large datasets and adherence to strict geometric constraints. 
In fixed-mounted PTZ camera setups, where the camera adjusts its view primarily through rotation and zooming rather than positional movement, the applicability and flexibility of traditional methods are limited. To achieve flexible image rectification for PTZ cameras, innovative works explored the implementation using sparse-view images\cite{zhang2020deepptz}. 
However, because of insufficient and unstable geometric information, these approaches exhibited limitations in rectification performance. 
Consequently, single-image rectification methods\cite{yang2023innovating} based on deep learning have gained significant attention. We also follow this idea.

\begin{figure}[t]
	\centerline{\includegraphics[width=\columnwidth]{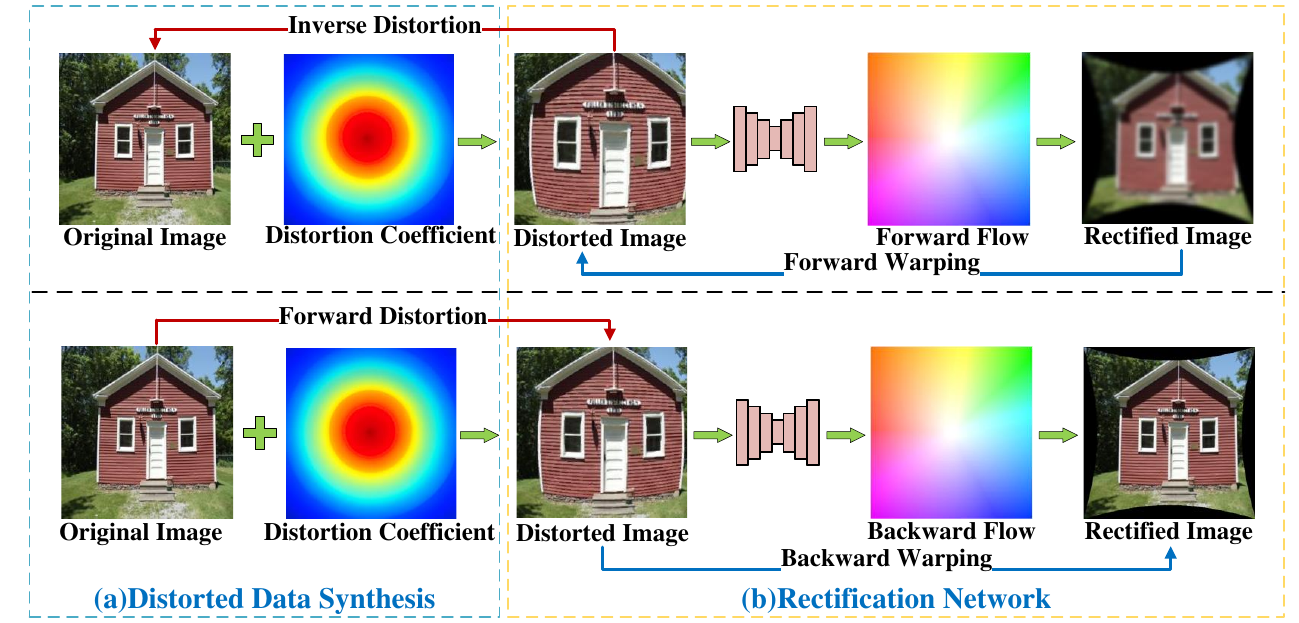}}
    \vspace{-10pt}

    \caption{This represents the two stages of the training process of image rectification. The top is the traditional pipeline, and the bottom is our method.}
    \vspace{-16pt}
	\label{image7-hou}
\end{figure}

The training process of these methods can be divided into two stages\cite{zhao2024model}: distorted data synthesis and rectification network (see Fig.\ref{image7-hou}). However, both are limited by the distortion strategy. In the stage of distorted data synthesis, distortion models are preferred for their ability to accurately fit complex lens distortion characteristics.
Traditional methods mostly use the inverse distortion model\cite{li2019blind,yang2021progressively}, which maps the distorted image back to the expected undistorted value. 
This is very natural, as it is similar to the process of image rectification. 
However, experiments indicate that employing this model during the data synthesis stage results in image blur and a loss of details.
Instead, we use a forward distortion model\cite{benligiray2015lens}, which applies distortion to camera rays made by projecting 3D points onto an image and calculates the offset directly. The resulting synthetic image has a very accurate pixel mapping, which helps to preserve image details during network training. 
In rectification network, these methods typically employ two approaches: parameter regression and image generation. The former employs deep neural networks to estimate distortion parameters\cite{liao2021deep, veicht2025geocalib}, 
while the latter directly generates rectified images in an end-to-end manner\cite{liao2019dr,yang2021progressively}. However, these methods still face great challenges in image detail restoration, such as non-integer pixel redundancy and artifacts.

To address existing challenges, we propose a Forward Distortion and Backward Warping Network (FDBW-Net). It begins by using a forward distortion model instead to synthesize barrel-distorted images to mitigate pixel redundancy and avoid image blur. The network leverages a pyramid context encoder to hierarchically extract and learn regional latent features. It incorporates a Backward Warping Estimation Module (BWEM), which applies channel-wise and spatial attention mechanisms to predict the precise backward warping flows for distortion rectification, enriched with geometric details.
Additionally, the multi-scale decoder employs a Layer-by-Layer Rectification Module (LLRM) to restore image details and output rectified images. For each distortion layer, the decoder progressively adjusts the distorted pixels using a backward warping strategy to ensure content consistency.
For evaluation, we validate the proposed FDBW-Net using public image datasets, synthetic PTZ camera imagery (we rendered them in AirSim\cite{shah2018airsim}), and real-scene PTZ camera datasets.
This paper makes three contributions:
\begin{itemize}
	\item We proposed FDBW-Net, a novel framework for wide-angle image rectification that enhances detail restoration by considering distortion strategies at both distorted data synthesis and rectification network.
	
	\item We used AirSim\cite{shah2018airsim} to render a set of image datasets from PTZ cameras at different perspectives and zooms to train the FDBW-Net, which built a bridge for application to PTZ cameras.
	
	\item Experiments demonstrate that FDBW-Net achieves SOTA performance in distortion rectification and has very good practicality in real-scene PTZ camera images. 
\end{itemize}

\section{Related Work}
\textbf{Distorted Data Synthesis.}
Various models have been developed to describe radial distortion.
Blind\cite{li2019blind} explored six models of geometric distortion (e.g., barrel, pincushion, and wave) to increase data diversity, hoping to fully represent the real distortion of the camera, but this greatly increases the complexity of the algorithm and is not flexible enough in practice.
Zhao \textit{et al.}\cite{zhao2024model} introduced a cascade model inspired by fisheye lenses, which combines multiple reversible distortion models into a unified framework. However, it did not specifically consider the perspective of PTZ cameras and was not verified in real-scene PTZ camera imagery.
The inverse distortion model used in  PCN\cite{yang2021progressively} is a popular barrel distortion model:
\begin{equation}
	\theta_u = \sum_{i=1}^n k_i \theta_d^{2i-1}, \quad (n=1,2,3,\ldots). 
	\label{eq:1forward_model}
\end{equation}
where \(\theta_u\) and \(\theta_d\) are the angles in undistorted and distorted lenses, with \(k_i\) as coefficients. 
Although this model was designed for image rectification, it often introduces blur and leads to a loss of detail in synthetic images.
To address this, we instead employ a forward distortion model, ensuring accurate pixel mapping from the undistorted image to the distorted space. This reduces artifacts and preserves details, making it highly effective for distortion rectification.

\textbf{Rectification Network }
The networks for image rectification can be broadly classified into two approaches: parameter regression and image generation.
In the former, Blind\cite{li2019blind} predicted forward warping flows to rectify perspective distortions in single images, but this often resulted in missing regions in the rectified image.
DeepCalib\cite{bogdan2018deepcalib} and Ordinal\cite{liao2021deep} estimated distortion parameters and focal lengths for effective rectification. However, they rely on limited  parameters and might lead to inaccuracies and rectification errors.
The recent RDTR\cite{wang2023model} employed a unified radial distortion model with distortion-aware pre-training to achieve robust geometric corrections, yet it struggled to maintain control over the rectified image quality.
Generative approaches were also investigated. DR-GAN\cite{liao2019dr} introduced the first adversarial network for radial distortion rectification, achieving pixel-level accuracy. However, this method sometimes produced blurred image content. 
PCN\cite{yang2021progressively} mitigated this issue by using a multi-scale loss function, which progressively refined distorted features, improving the robustness and prediction accuracy. Nonetheless, its self-supervised approach faces challenges in generating accurate appearance flows.
In contrast, QueryCDR\cite{guo2024querycdr} employed a distortion-aware learnable query mechanism to enhance the rectification of various distortions, but it responds slowly to local features, resulting in the loss of output details.

\begin{figure*}[htbp]

	\centering
	\includegraphics[width=\linewidth]{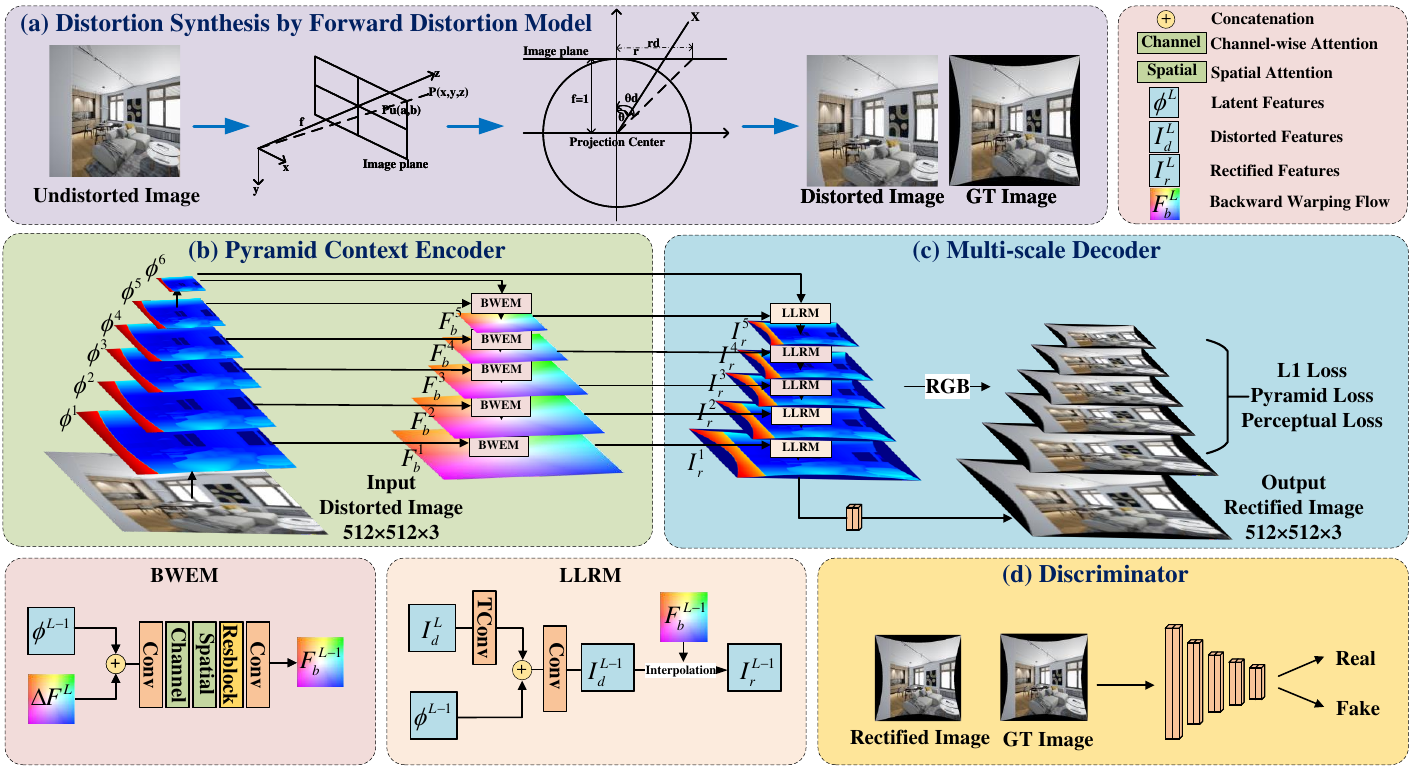}
    \vspace{-20pt}
	\caption{The overall structure of our FDBW-Net. ``BWEM" means the backward warping estimation module and ``LLRM" means the layer-by-layer rectification module. In discriminator, ``Real" means the ground truth images and ``Fake" means the images generated by the generator.}
    \vspace{-15pt}
	\label{fig:1network}
\end{figure*}

\section{Distorted Data Synthesis of FDBW-Net}
\label{sec:Distortion models}

\label{sec:Synthetically distorted data}

In the following, we provide a detailed introduction to a forward distortion model and then adapt it to our work. For a 2D point \(P\) in the original distortion-free image and assume that the projection matrix \(K\) is known.
\[
K = \begin{bmatrix}
	f & 0 & c_x \\
	0 & f & c_y \\
	0 & 0 & 1
\end{bmatrix}
\]

In Fig.\ref{fig:1network} (a), it is converted to \(P_u(a,b)\) in the camera coordinate system, which establishes a projective geometry with the 3D point \(P_{(x,y,z)}\). 
Note that we normalize the focal length to unity, that is, \(f=1\), thus mitigating the influence of the focal length in subsequent distortion synthesis. 
The distance \(r_u\) is from the center of the image to \(P_0(a_0, b_0)\), and the angle \(\theta_u\) represents the intersection with the optical axis. The perspective ray is determined by \eqref{eq:2calculate_angle}:
\begin{equation}
	\theta_u = \arctan\left(\frac{r_u}{f}\right) = \arctan(r_u)
	\label{eq:2calculate_angle}
\end{equation}
Then, the forward distortion model \eqref{eq:3reverse_model} is applied to \(\theta_u\) to obtain \(\theta_d\) . 
\begin{equation}
	\theta_d = \theta_u \left(1 + k_1 \theta_u^2 + k_2 \theta_u^4 + k_3 \theta_u^6 + k_4 \theta_u^8 \right)
	\label{eq:3reverse_model}
\end{equation}
where \(k_1, k_2, k_3, k_4\) represent distortion coefficients.

Consequently, we obtain the corresponding distorted point \(P_d(u,v)\) from the original image point \(P\). By this method, we synthesize the distorted image and its corresponding ground truth (GT) image.

\section{Rectification network of FDBW-Net}
\label{sec:approach}

As shown in Fig.\ref{fig:1network}, the network of FDBW-Net consists of three main components: a pyramid context encoder, a multi-scale decoder, and a discriminator. 

\subsection{Pyramid Context Encoder}
\label{sec:Encoder}

\textbf{Pyramid context extractor.} 
We employ a pyramid context extractor that learns latent features at six layers. These features capture key geometric structures and textures from the distorted image, with the lower layers focusing on local details and the higher layers capturing global patterns and structural relationships. Specifically, in the pyramid context encoder with \( L = 6 \) layers, the distorted image \( \mathbf{I_d} \in \mathbb{R}^{h \times w \times 3} \) is passed through six layers of convolution. Each layer uses a \( 3 \times 3 \) kernel, stride 2, batch normalization, and ReLU activation. 
The resolution of the latent features progressively decreases to 256, 128, 64, 32, 16, and 8. The latent features are denoted as \( \phi_L, \phi_{L-1}, \dots, \phi_1 \), as shown in Fig.\ref{fig:1network} (b).

\textbf{Backward Warping Estimation.} 
We design a backward warping estimation module (BWEM) to generate warping flows representing the pixel offsets of distortion. To preserve the geometric details of the flows, we first use channel-wise attention to compress the potential features and dynamically adjust the importance in each channel to focus on the key features that contain global information. Then, we use spatial attention to assign higher weights to the features of important regions in order to estimate accurate offsets for distortion areas. Finally, the residual block outputs warping flows. 

For each layer \( L \) of the pyramid, BWEM generates a warping flow \( F_{b}^{L} \). By iteratively applying BWEM across multiple layers, it combines the global features in the higher layers and the features in the lower layers with the fine-grained details, as shwon in Fig. \ref{fig:1network}. For each iteration, it generates refined offsets \( \Delta F^{L} \) learned through convolutional layers for the warping flows. These flows are computed as:
\begin{equation}
	\begin{aligned}
		F_{b}^{L-1} &= \textbf{BWEM}(\phi^{L-1}, \Delta F^{L})
	\end{aligned}
	\label{eq:BEM}
\end{equation}

\subsection{Multi-scale Decoder}
\label{sec:Decoder}

We use the multi-scale decoder to deal with features and generate the rectified images together with a layer-by-layer rectification module (LLRM). In each layer of LLRM (see Fig.\ref{fig:1network}), the distorted features \(I_{d}^{L}\) are first upsampled by transposed convolutions to preserve the global image structure and then progressively fused with the latent features \(\phi^{L-1}\) . Then, the warping flow \( F_{b}^{L} \) is fed to rectify the distortion offset with a backward bilinear interpolation. As a result, it obtains the coordinates of the rectified features of each layer. As shown in \eqref{eq:6bilinear sampling}, \(I_r^L(u, v)\) represents the coordinates of the rectified features and \((u + F^L_{bx}(u), v + F^L_{by}(v))\) corresponds to the mapped coordinates of the distorted features. This similar operation is carried out progressively between different layers to ensure the restoration of geometric details.
\begin{equation}
    I_r^L(u, v) = I_d^L\left(u + F^L_{bx}(u), v + F^L_{by}(v)\right),
    \label{eq:6bilinear sampling}
\end{equation}

The rectified features are passed through a 3x3 convolutional layer to generate multi-scale rectified images \( \mathbf{I_r^L} \quad (L = 1, \dots, 5) \). In the final layer, a transposed convolution enlarges the features to produce the high-resolution rectified image \( \mathbf{I_r} \in \mathbb{R}^{h \times w \times 3} \). As a result, the multi-scale decoder processes the warping flows in a backward way to  rectify the image.

\subsection{Discriminator}

The discriminator is used to effectively differentiate between the ground truth images and the images generated by the generator.
It consists of six \( 5 \times 5 \) convolutional layers with a stride of 2, configured with 64, 128, 256, 512, 512, and 512 filters. Each layer incorporates Batch Normalization and LeakyReLU activation to ensure stable and efficient training.
After the convolution, two fully connected layers are passed to output the final feature map.

\vspace{-3mm}
\subsection{Loss functions }
\label{sec:Loss function}
 The generator is trained with a combination of L1 loss, perceptual loss, pyramid loss, and adversarial loss. The discriminator is optimized using adversarial loss.

We form an integrated loss function \(L\) for our FDBW-Net to ensure precise and detailed image rectification. 
\begin{equation}
	\begin{aligned}
		L = \lambda_1 L_{\text{L1}} + \lambda_2 L_{\text{perc}} + L_{\text{pyramid}} + L_{\text{adv}}
	\end{aligned}
	\label{eq:10all}
\end{equation}
Where, \(\lambda_1, \lambda_2\) are hyperparameters used to adjust the weights of the different loss functions.

Specifically, the L1 loss \(L_{L1}\) ensures structural similarity between the rectified image \(I_r\) and the ground truth image \(I_{gt}\). The perceptual loss \(L_{\text{prec}}\) leverages features extracted from the pre-trained VGG19 network to preserve high-level structural details. Pyramid loss \(L_{\text{pyramid}}\) captures image details across multiple scales by computing losses at progressively lower resolutions. Lastly, the adversarial loss \(L_{\text{adv}}\), optimized through the interaction between the generator and the discriminator, promotes realistic textures, encouraging the generation of images that are indistinguishable from the real ones. These loss functions are defined as follows:

\begin{equation}
	\begin{aligned}
		L_{L1} = \left\lVert I_{r} - I_{gt} \right\rVert_1
	\end{aligned}
	\label{eq:7Lr}
\end{equation}
\vspace{-3mm}
\begin{equation}
	\begin{aligned}
		L_{\text{prec}} = \sum_{i=1}^{5}  w_i \left\lVert K_i^{\text{VGG}}(I_{r}) -K_i^{\text{VGG}}(I_{gt}) \right\rVert_1
	\end{aligned}
	\label{eq:8Lprec}
\end{equation}
\vspace{-3mm}
\begin{equation}
	\begin{aligned}
		L_{\text{pyramid}} = \sum_{i=1}^{5}  \left\lVert I_r^i - \text{interp}(I_{\text{gt}}, \text{size}(I_r^i)) \right\rVert_1
	\end{aligned}
	\label{eq:9Lpyra}
\end{equation}
\vspace{-3mm}
\begin{equation}
    L_{\text{adv}} = \min_{G} \max_{D} \left( 
    \mathbb{E} \left[ \log D(I_{\text{gt}}) \right] + 
    \mathbb{E} \left[ \log \left( 1 - D(G(I_{r})) \right) \right] 
    \right)
\end{equation}
Where, \(K_i^{\text{VGG}}\) is the graph of features extracted by VGG19 network, \(w_i\) and is the weight of feature loss at each level.
\(I_{gt}\) is scaled to the same size as \(I_{r}\).
Minimizing the generator \(G\) and maximizing the discriminator \(D\).


\section{Experiments}
\label{sec:experiments}

We quantitatively and qualitatively compare FDBW-Net (denoted as Ours) with various image rectification methods, including DeepCalib \cite{bogdan2018deepcalib}, Blind\cite{li2019blind}, Ordinal \cite{liao2021deep}, RDTR\cite{wang2023model}, DR-GAN \cite{liao2019dr}, PCN \cite{yang2021progressively} and QueryCDR\cite{guo2024querycdr}. And Ours outperforms them in multiple metrics.

\begin{table}[htbp]

	\caption{Comparison of various methods on the Places365 dataset\cite{zhou2017places}.}
    \vspace{-20pt}
	\begin{center}
		\begin{tabular}{|c|c|c|c|c|c|c|}
			\hline
			\textbf{Type} & \textbf{Name}  & \textbf{PSNR↑} & \textbf{SSIM↑} & \textbf{EPE↓} & \textbf{FID↓} \\ 
			\hline
			\multirow{4}{*}{\begin{tabular}[c]{@{}c@{}}Parameter \\ Regression\end{tabular}} 
			& DeepCalib\cite{bogdan2018deepcalib} & 17.57 & 0.53 & 9.79 & 14.26 \\ 
			\cline{2-6} 
			& Blind\cite{li2019blind}  & 9.01 & 0.38 & 15.17 & 203.94  \\ 
			\cline{2-6} 
			& Ordinal\cite{liao2021deep} & 25.07 & 0.88 & 10.23 & 18.02 \\ 
			\cline{2-6} 
			& RDTR\cite{wang2023model}  & 30.35 & 0.93 & 11.54 & 55.87 \\ 
			\hline
			\multirow{3}{*}{\begin{tabular}[c]{@{}c@{}}Image \\ Generation\end{tabular}} 
			& DR-GAN\cite{liao2019dr} & 21.26 & 0.68 & 8.42 & 9.84 \\ 
			\cline{2-6} 
			& PCN\cite{yang2021progressively}  & 28.81 & 0.90 & 5.30 & 4.09 \\ 
			\cline{2-6} 
                & QueryCDR\cite{guo2024querycdr}  & 29.79 & 0.91 & 5.24 & 3.05\\ 
                \cline{2-6} 
			& FDBW-Net(Ours)  & \textbf{31.70} & \textbf{0.95} & \textbf{4.38} & \textbf{2.81} \\ 
			\hline
			\multicolumn{5}{l}{$^{\mathrm{a}}$↑ Higher is better, ↓ Lower is better}
		\end{tabular}
		\label{tab:1comparisons}
        \vspace{-15pt}
	\end{center}

\end{table}
\begin{figure}[htbp]
	\centerline{\includegraphics[width=\columnwidth]{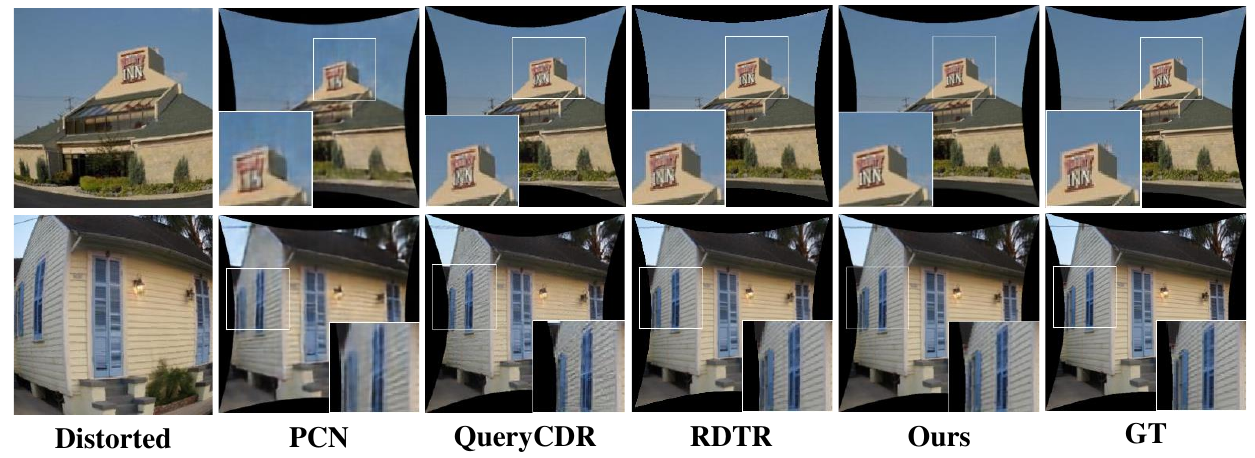}}
        \vspace{-10pt}
    \caption{Comparison in detail recovery of PCN\cite{yang2021progressively}, QueryCDR\cite{guo2024querycdr}, RDTR \cite{wang2023model} and Ours.}
        \vspace{-10pt}
	\label{image8-duibi}
\end{figure}

\subsection{Implementation details}
\label{sec:Implementation details}

\textbf{Evaluation Metrics.} 
We use PSNR and SSIM\cite{bogdan2018deepcalib} to evaluate the difference between the rectified image and the ground truth, measuring pixel-level accuracy and structural fidelity. Additionally, we introduce FID and EPE metrics\cite{yang2023innovating} to assess the feature-level differences in high-dimensional space.

\textbf{Training Configuration.} Hyperparameters \(\lambda_1,\lambda_2\)  are set to 120 and 10, respectively. The model is optimized using the Adam optimizer with a learning rate of \(10^{-4}\), and training is conducted on an NVIDIA GeForce RTX 4090D GPU.

\subsection{Comparative Analysis}
\label{sec:Comparative Analysis}

\textbf{Quantitative Comparison.}
Trainable models are trained on the Places365 dataset \cite{zhou2017places} with 30,000 images and evaluated on a test set of 3,000 images, which includes ground truth images and synthetic barrel-distorted ones generated using the forward distortion model. For Blind \cite{li2019blind} and RDTR \cite{wang2023model}, we evaluate their pre-trained models on our test dataset.
 
As shown in Table \ref{tab:1comparisons}, Ours excels in all evaluation metrics. DeepCalib, Blind, and Ordinal rely on limited parameter estimations, limiting their ability, especially at image edges. For example, Blind often produces noticeable missing regions. 
DR-GAN suffers from encoder overload, which leads to a lack of smooth context and results in images with blur. Ordinal is constrained by inherent distortion patterns. 
While PCN, QueryCDR and RDTR perform well in certain aspects, they still struggle with fine-grained detail recovery.

\textbf{Qualitative Comparison.}
To provide an intuitive comparison of the rectification results, we visualize the outputs of PCN, RDTR, and Ours on the Places365 dataset, as shown in Fig.\ref{image8-duibi}.
PCN struggles to preserve complex textures and fine details, leading to imprecise rectified images.
QueryCDR's dynamic control adjustments respond slowly to local information when focusing on different features, resulting in limited restoration of output details.
While RDTR captures more intricate details through perceptual pre-training, it often introduces edge artifacts, such as inconsistent distortions, blurred boundaries, and jagged edges, which degrade overall visual quality, especially around object contours.
In contrast, Ours excels at preserving fine details while minimizing information loss and blur. By extracting geometric features via an encoder and refining them with backward warping flows, our method fully preserves key content features. This allows the decoder to effectively reconstruct finer details, achieving superior structural accuracy and visual fidelity compared to existing methods.

\subsection{Experiments on PTZ cameras}
\label{sec:PTZ verification}

\textbf{ Rendered PTZ Camera Imagery.}
We render PTZ camera images using AirSim at different perspectives and zooms, resulting in a dataset of 17,000 original images. Using the forward distortion model, we generate distorted images and their corresponding ground truth to train our FDBW-Net. For comparison, we evaluate FDBW-Net against DR-GAN \cite{liao2019dr}, Ordinal \cite{liao2021deep}, PCN \cite{yang2021progressively}, and QueryCDR\cite{guo2024querycdr}.

\begin{figure}[htbp]
    \vspace{-5pt}
	\centerline{\includegraphics[width=\columnwidth]{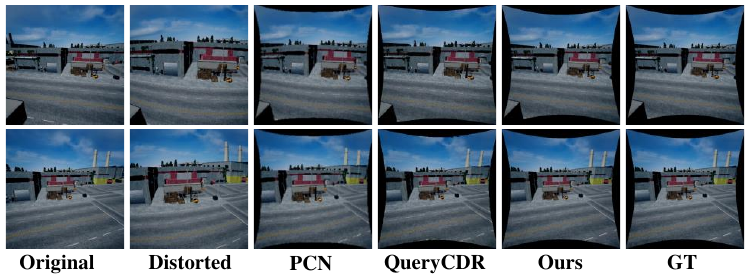}}
    \vspace{-8pt}
	\caption{Visualization of synthetic PTZ camera images from various views.}
    \vspace{-10pt}
	\label{image5-synthesizePTZ}
\end{figure}

Table \ref{tab:4comparison PTZ} summarizes the results, clearly showing that FDBW-Net excels in addressing wide-angle distortion rectification for PTZ camera viewpoints, consistently outperforming all the methods compared in various performance metrics. Additionally, Fig.\ref{image5-synthesizePTZ} provides a visual comparison between images rectified by PCN, QueryCDR and our method. The results demonstrate that FDBW-Net significantly surpasses PCN and QueryCDR in both detail preservation and overall image quality. The images estimated by FDBW-Net are much closer to real PTZ view scenes, further validating its superiority and strong generalization capabilities.

\begin{figure}[htbp]
	\centerline{\includegraphics[width=\columnwidth]{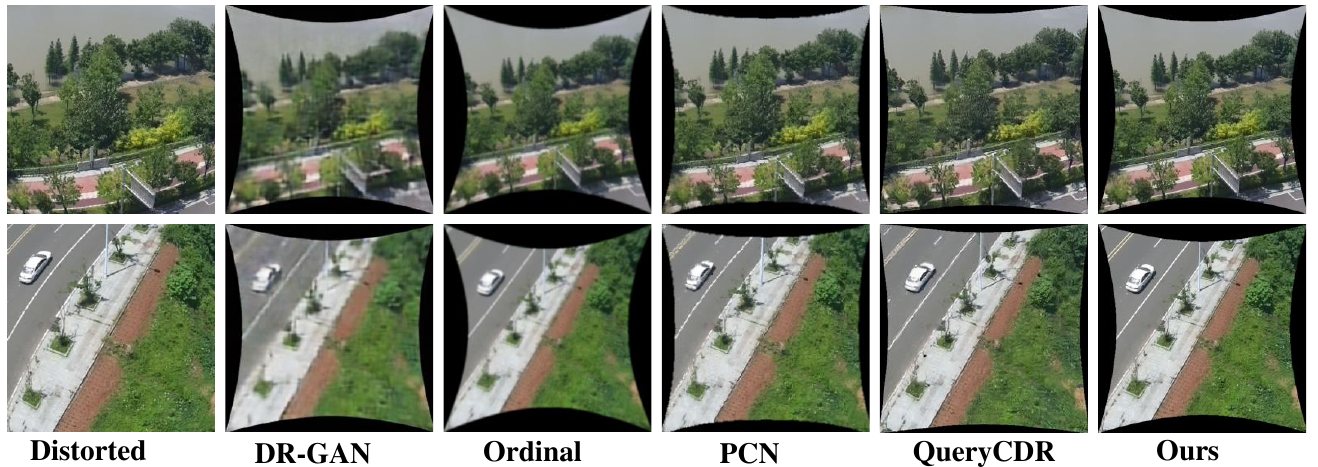}}
    \vspace{-8pt}
	\caption{Visualization of real-scene images from PTZ cameras.}
	\label{image6-realworldPTZ}
    \vspace{-10pt}
\end{figure}

\textbf{Real-scene PTZ Images.}
To verify the generalization ability of FDBW-Net, we conduct additional experiments on distorted images of real-scene PTZ cameras. We capture images using wide-angle lenses with varying focal lengths and fields of view and then directly predict these images using our FDBW-Net with the model weights trained from the previous rendered PTZ images. The columns in Fig.\ref{image6-realworldPTZ} represent different perspectives of the PTZ camera. 
Despite the inherent differences between estimated wide-angle images and real-scene images, the experimental results indicate that FDBW-Net consistently produces rectified images with high detailed quality. 
This further validates the practical applicability of FDBW-Net in real-world scenarios.

\begin{table}[htbp]
        \vspace{-5pt}
	\caption{Numerical Analysis of Methods on Synthesized PTZ Images}
        \vspace{-10pt}
	\begin{center}
		\begin{tabular}{|c|c|c|c|c|}
			\hline
			\textbf{Method} & \textbf{PSNR↑} & \textbf{SSIM↑} & \textbf{EPE↓} & \textbf{FID↓} \\ 

			\hline
			DR-GAN\cite{liao2019dr} & 21.27 & 0.68 & 10.23 & 18.02 \\
			\hline
			Ordinal\cite{liao2021deep} & 15.17 & 0.36 & 11.54 & 55.88 \\
			\hline
			PCN\cite{yang2021progressively} & 21.14 & 0.73 & 6.55 & 12.98 \\
            \hline
			QueryCDR\cite{guo2024querycdr} &25.99 & 0.85 & 8.59 & 9.91 \\
			\hline
			FDBW-Net(Ours) & \textbf{30.35} & \textbf{0.93} & \textbf{5.08} & \textbf{3.71} \\
			\hline
		\end{tabular}
		\label{tab:4comparison PTZ}
                \vspace{-15pt}
	\end{center}
\end{table}

\subsection{Ablation Study}
\label{sec:Ablation Study}

\textbf{Ablation of Distorted Data Synthesis.}
To evaluate the impact of our distortion data synthesis method, we conduct a focused comparison with PCN\cite{yang2021progressively}, separating the distortion data synthesis process from the network architecture. In this setup, the distortion data synthesis is denoted as \textit{S}, and the network architecture as \textit{N}. We train and
test on the Places365\cite{zhou2017places} dataset, and the experimental results are summarized in Table \ref{tab2:synthetic_data_comparison}.
For the PCN network, the use of the forward distortion model to synthesize the data yields better results. Despite introducing higher distortion coefficients and more pronounced distortion values, this approach effectively reduces interpolation artifacts and preserves finer image details, thereby demonstrating the effectiveness of our forward distortion model.
Furthermore, applying PCN’s data synthesis method to our FDBW-Net architecture also produces superior rectification performance, which further validates the effectiveness of our pyramid architecture.

\begin{table}[htbp]
    \vspace{-10pt}
    \caption{Comparison of Synthetic Data Methods}
    \vspace{-10pt}
    \begin{center}
        \begin{tabular}{|c|c|c|c|c|c|}
            \hline
            \textbf{Data Synthetic} & \textbf{Network} & \textbf{PSNR↑} & \textbf{SSIM↑} & \textbf{EPE↓} & \textbf{FID↓} \\
            \hline
            \(S_{\text{PCN}}\) & \(N_{\text{PCN}}\) & 27.14 & 0.83 & 8.02 & 13.77 \\
            \hline
            \(S_{\text{FDBW\_Net}}\) & \(N_{\text{PCN}}\) & 28.81 & 0.90 & 5.30 & 4.09 \\
            \hline
            \(S_{\text{PCN}}\) & \(N_{\text{FDBW\_Net}}\) & 30.80 & 0.93 & 5.38 & 4.15 \\
            \hline
            \(S_{\text{FDBW\_Net}}\) & \(N_{\text{FDBW\_Net}}\) & \textbf{31.70} & \textbf{0.95} & \textbf{4.38} & \textbf{2.81} \\
            \hline
        \end{tabular}
        \label{tab2:synthetic_data_comparison}
    \end{center}
    \vspace{-10pt}
\end{table}

\begin{table}[htbp]
    \vspace{-10pt}
    \caption{Ablation Study: Impact of BWEM and LLRM Removal.}
    \vspace{-5pt}
    \centering
    \begin{tabular}{|l|c|c|c|c|}
        \hline
        \textbf{Configuration} & \textbf{PSNR}↑ & \textbf{SSIM}↑ & \textbf{EPE}↓ & \textbf{FID}↓ \\
        \hline
        w/o BWEM & 18.22 & 0.54 & 10.52 & 219.91 \\
        \hline
        w/o LLRM & 23.46 & 0.73 & 8.57 & 35.62 \\
        \hline
        Full Model & \textbf{30.34} & \textbf{0.93} & \textbf{5.07} & \textbf{3.70} \\
        \hline
    \end{tabular}
    \vspace{-10pt}
    \label{tab:2ablation Flow}
\end{table}

\begin{figure}[htbp]
	\centerline{\includegraphics[width=\columnwidth]{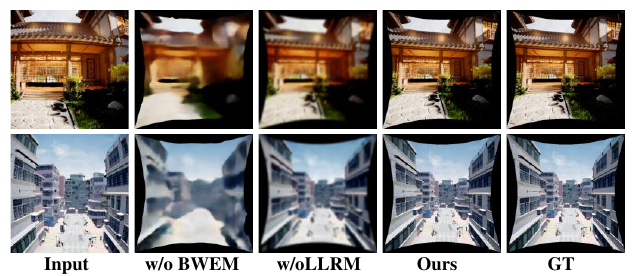}}
    \vspace{-10pt}
	\caption{Visualization of BWEM and LLRM Removal.}
    \vspace{-5pt}
	\label{image4-ablationimage}
\end{figure}

\textbf{Ablation of Network Modules.}
To quantitatively assess the contribution of different modules, we perform ablation experiments using AirSim-rendered PTZ images with the forward distortion model. We remove the Backward Warping Estimation Module from the network (denoted as w/o BWEM) and remove the Layer-by-Layer Rectification Module (denoted as w/o LLRM).

The results are presented in Table \ref{tab:2ablation Flow} and Fig.\ref{image4-ablationimage}. As observed, the removal of the BWEM significantly degrades rectification performance, leading to a noticeable loss of content. 
Specifically, the absence of this module results in poor preservation of fine-grained features and complex textures, leading to blurred and incomplete reconstructions.
This demonstrates that the backward warping strategy plays a crucial role as it enables the model to handle different degrees of distortion in different image regions.
Moreover, removing the LLRM results in only the backward warping flow from the largest layer being used to map the distorted image, which leads to distorted object boundaries, misaligned contours, and unnatural edge artifacts. When LLRM is included, overall visual fidelity is greatly improved, and the geometric alignment of image features becomes more accurate.

These findings underscore the complementary roles of BWEM and LLRM: BWEM is essential for preserving geometric features, while LLRM is crucial for progressively refining the recovery of details across layers. The ablation study reinforces the critical importance of these modules within the full FDBW-Net architecture for addressing wide-angle image rectification challenges.


\textbf{Loss Ablation.}
\label{sec:Loss Ablation}
Because L1 loss and adversarial loss are often primary and necessary, we investigate the effect of adding perceptual loss \(L_{\text{prec}}\) and pyramid loss \(L_{\text{pyramid}}\), and the results are shown in Table \ref{tab:3loss_ablation}. 
When perceptual loss or pyramid loss is added, network performance improves significantly. Perceptual loss enhances overall image quality by focusing on high-level features and improving structural consistency, while pyramid loss refines details through multi-level feature fusion. In particular, the best results are achieved when both losses are combined.

\begin{table}[htbp]
    \vspace{-10pt}
	\caption{
    Ablation Study: Impact of Perceptual and Pyramid Loss}
    \vspace{-10pt}
	\begin{center}
		\begin{tabular}{|c|c|c|c|c|c|}
			\hline
			\textbf{\(L_{\text{prec}}\)}  & \textbf{\(L_{\text{pyramid}}\)}  & \textbf{PSNR↑} & \textbf{SSIM↑} & \textbf{EPE↓} & \textbf{FID↓} \\
			\hline
			— & — & 28.24 & 0.89 & 5.96 & 8.29 \\
			\hline
			\checkmark & — & 29.32 & 0.91 & 5.48 & 4.97 \\
			\hline
			— & \checkmark & 29.39 & 0.91 & 5.44 & 5.15 \\
			\hline
			\checkmark & \checkmark & \textbf{30.34} & \textbf{0.93} & \textbf{5.07} & \textbf{3.70} \\
			\hline
		\end{tabular}
		\label{tab:3loss_ablation}
	\end{center}
    \vspace{-20pt}
\end{table}

\section{Conclusion}
\label{sec:conclusion}
In this paper, we present FDBW-Net, a novel deep learning-based approach for single image rectification in PTZ camera settings. It leverages a forward distortion model to synthesize training data and employs predicted backward warping flows to progressively rectify distorted images. It obtains significant improvements in both the accuracy of rectification and the preservation of fine details. Extensive experiments validate that FDBW-Net effectively addresses the limitations of traditional multi-view geometry-based methods, offering enhanced flexibility in PTZ camera image rectification. Furthermore, the results demonstrate that FDBW-Net is well-suited for practical deployment, showing considerable promise in real-world vision-based PTZ applications.


\bibliographystyle{IEEEbib}
\bibliography{icme2025references}

\end{document}